\documentclass[conference]{IEEEtran}

\usepackage{times}
\usepackage{epsfig}
\usepackage{graphicx}
\usepackage{amsmath}
\usepackage{amssymb}
\usepackage[dvipsnames]{xcolor}
\usepackage[numbers,sort,compress]{natbib}
\usepackage{titletoc}
\usepackage{algpseudocode}
\usepackage{algorithm}
\usepackage{caption}
\usepackage{subcaption}
\usepackage{textcomp}
\usepackage{booktabs}
\usepackage{amsmath}
\usepackage{hyperref}
\hypersetup{
 colorlinks,
 citecolor=blue,
 linkcolor=Red,
 urlcolor=violet}
\usepackage[english]{babel}
\usepackage{tikz}
\usetikzlibrary{calc}
\usepackage{lipsum}

\begin{document}

\IEEEoverridecommandlockouts
\IEEEpubid{\makebox[\columnwidth]{Peer reviewed \& accepted at  \href{http://www.vcip2022.org/}{IEEE VCIP 2022} \hfill} \hspace{\columnsep}\makebox[\columnwidth]{ }}

\title{MAiVAR: Multimodal Audio-Image and Video Action Recognizer}
\author{
 Muhammad Bilal Shaikh\textsuperscript{1}, Douglas Chai\textsuperscript{1}, Syed Mohammed Shamsul Islam\textsuperscript{1} and Naveed Akhtar\textsuperscript{2}\\
 \textsuperscript{1}Edith Cowan University, 270 Joondalup Drive, Joondalup, WA 6027, Perth, Australia\\ \textsuperscript{2}The University of Western Australia, 35 Stirling Highway, Crawley, WA 6009, Perth, Australia  \\
 \textsuperscript{1}\{m.shaikh,d.chai,syed.islam\}@ecu.edu.au, \textsuperscript{2}naveed.akhtar@uwa.edu.au} 

\renewcommand{\algorithmicrequire}{\textbf{Input:}}
\renewcommand{\algorithmicensure}{\textbf{Output:}}

\newcommand{\Bilal}[1]{\textcolor{red}{#1}}
\newcommand{\Shams}[1]{\textcolor{green}{#1}}
% Add author with custom color custom colors

\newcommand{\Douglas}[1]{\textcolor{blue}{#1}}
\newcommand{\Naveed}[1]{\textcolor{orange}{#1}}

\newcommand{\comment}[1]{}

\maketitle
%\tikz[overlay,remember picture]
%{
    %\node at ($(current page.west)+(1,0)$) [rotate=90] %{\Huge\textcolor{gray}{arXiv:2207.10258v1 [cs.CV]: \today}};
%}
\begin{abstract}
%We did something awesome and wrote about it.

Currently, action recognition is predominately performed on video data as processed by CNNs. We investigate if the representation process of CNNs can also be leveraged for multimodal action recognition by incorporating image-based audio representations of actions in a task.
To this end, we propose Multimodal Audio-Image and Video Action Recognizer (MAiVAR),  a  CNN-based audio-image to video fusion model that accounts for video and audio modalities to achieve superior action recognition performance.
MAiVAR extracts meaningful image representations of audio and fuses it with video representation to achieve better performance as compared to both modalities individually on a large-scale action recognition dataset. 

\end{abstract}

%%%%%%%%% BODY TEXT

\section{Introduction}
Action recognition techniques proposed in the last decade have predominantly used video representations, i.e., sequences of images to recognize an event or action in a video \cite{shaikh2021rgb,li2019collaborative,wu2018compressed}. Visual modality contains spatial information, which is inherently helpful for CNN-based classiﬁcation. To better capture the multimodality aspect of action data, a recent trend has been to combine information from different modalities such as optical ﬂow, RGB-difference, warped-optical ﬂow and others. These types of hybrid features can be applied to various action recognition techniques such as Temporal Segment Networks (TSN) \cite{wang2016temporal}, Temporal Relation Network (TRN) \cite{zhou2018temporal} and Temporal Shift Module (TSM) \cite{lin2019tsm}. Motivated by the success of
CNN-based models in object detection \cite{ren2015faster} and image classiﬁcation \cite{russakovsky2015imagenet}, this paper investigates whether CNNs have the ability to understand complex image-based audio representations and to impact multimodal classification. \par
\begin{figure}
    \centering
    \begin{subfigure}[b]{0.48\textwidth}
         \centering
         \includegraphics[width=\textwidth]{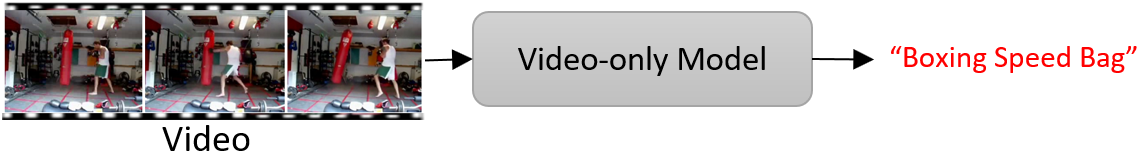}
         \setlength{\belowcaptionskip}{-2pt}   
         \end{subfigure}
     \hfill
     \begin{subfigure}[b]{0.48\textwidth}
         \centering
         \includegraphics[width=\textwidth]{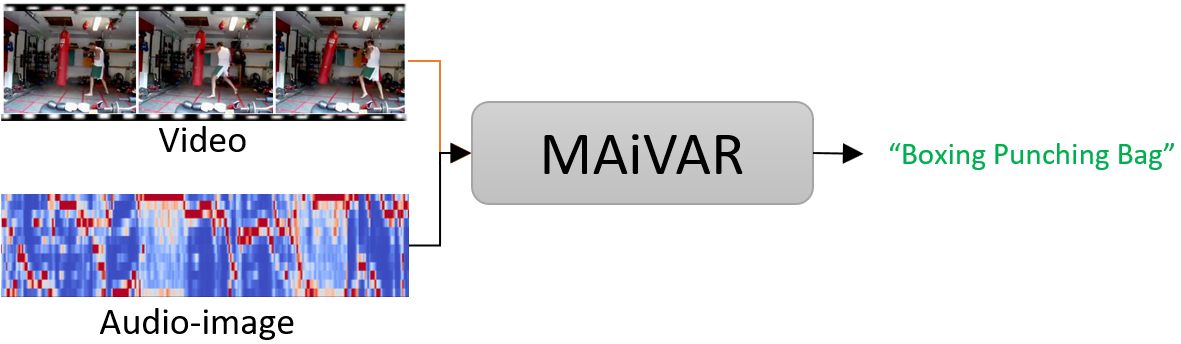}
     \end{subfigure}
     \setlength{\belowcaptionskip}{-8pt}   
    \caption{Action recognition using unimodal video-only approach (top) versus \textbf{MAiVAR} with intermediate fusion of audio-visual features (bottom).}
    \label{fig:teaser}
\end{figure}

The audio modality has significant features that can contribute significantly in action recognition from videos: audio contains dynamics and rich contextual temporal information \cite{gaver1993world} and requires a lighter computational process as compared to video frames. For example, as shown in Fig.~\ref{fig:teaser}, within a short clip of action of Boxing a Punching Bag, a single audio-image frame includes most of the dynamic contextual information contained in the audio, i.e., { the sound of a boxing glove hitting the bag}, while the accompanying video clip contains useful cues of spatial dynamics. This prompts the question of: across an entire video, can audio be beneficial to select the critical features that are useful for recognizing actions? 
To answer this question, we introduce the Multimodal Audio-Image and Video Action Recognizer (MAiVAR), a CNN-based audio-image to video fusion-based model that can be applied to video and audio features for classifying actions at a better accuracy than the unimodal equivalent. Additionally, we propose an approach to extract meaningful image representations of audio, which can significantly improve classification scores when used with CNNs. \par
The benefits of MAiVAR are threefold. Firstly, MAiVAR has superior performance, outperforming the state-of-the-art model on the same data for video representation when evaluated based on the UCF-101 dataset. Secondly, MAiVAR naturally supports both audio and video inputs and can be applied to different tasks without any architectural change. Thirdly, as compared to state-of-the-art video-based CNN models, MAiVAR features a simpler and more intuitive architecture. To the best of our knowledge, MAiVAR is the first audio-image to video fusion-based action classification model. The key contributions of this work are as follows: (1) a new feature representation strategy is proposed to select the most informative candidate representations for audio-visual fusion; (2) collection of effective audio-image based representations that complements video modality for better action recognition are included; (3) a novel fusion architecture, MAiVAR is proposed for audio-visual fusion that supports different audio-image representations and can be applied to different tasks; and (4) state-of-the-art results for action recognition on audio-visual dataset have been reported. 
\section{Related Work}
\label{sec:related}
Multimodal action recognition employs multi-stream approaches to incorporate different modalities. TSN \cite{wang2016temporal}, TRN \cite{zhou2018temporal} and TSM \cite{lin2019tsm} are based on 2D CNNs. All three models make use of a two-stream approach that uses both RGB and optical flow. Besides RGB and optical flow streams, Temporal Binding Networks (TBN) \cite{kazakos2019epicfusion} adds audio as another modality. SlowFast \cite{feichtenhofer2019slowfast} uses two RGB streams with different resolutions and frame rates. The proposed MAiVAR is based on TSN \cite{wang2016temporal} pre-trained on the Kinetics \cite{carreira2017quo} dataset and Inception-ResNet2 (IRv2) pre-trained on ImageNet \cite{russakovsky2015imagenet}. TSN was primarily proposed for video-level action recognition tasks, while IRV2 is a CNN architecture that builds on the inception family \cite{szegedy2016inceptionv4} of architectures but incorporates residual connections.\par 
Recently, TSN has been adapted for the backbone module in video understanding scenarios \cite{Lei_2021_CVPR,girdhar2017actionvlad,li2016vlad3,zhou2018temporal,Kwon_2021_ICCV}, where it is typically used in conjunction with a succeeding module. In \cite{Kwon_2021_ICCV}, TSN was employed as 2D CNN backbones to learn motion dynamics in videos. In contrast, IRV2 has been used for feature extraction from images in \cite{mei2020artificial}, which helped in different image restoration tasks \cite{gu2021ntire, yan2021precise}. Spatial-Temporal Network (STNeT) \cite{he2019stnet} adapts IRV2 to model local and global spatio-temporal features. The closest works to ours are \cite{wang2016temporal,takahashi2017aenet}. Takahashi \textit{et al.} have classified audio events using 3D CNN with some representations of audio, while the TSN-based approach \cite{wang2016temporal} uses optical flow variants with RGB. \par 
In another work \cite{gao2020listentolook}, a video skimming mechanism is applied for untrimmed video aided by audio to eliminate both short-term and long-term redundancies. IMGAUD2VID uses audio to extract dynamic scene information in a single frame, which captures most of the appearance information to form an audio-image pair. These pairs are then used to select key moments from the video for action recognition. Unlike IMGAUD2VID, our idea capture more spatial information along with the holistic dynamic information of a scene from the image representations of audio. \par 
The proposed MAiVAR differs from these works in that it extracts audio features using IRV2 and video features using TSN, fusing both features using a multi-layer perceptron (MLP), which is initialized with weights from an MLP  pre-trained with video features. We also conduct extensive experiments to show the design choice of MAiVAR on audio-image representations and fusion tasks. 
\section{Methodology}
\label{sec:method}
\begin{figure}[t]
     \centering
     \includegraphics[width=0.5\textwidth]{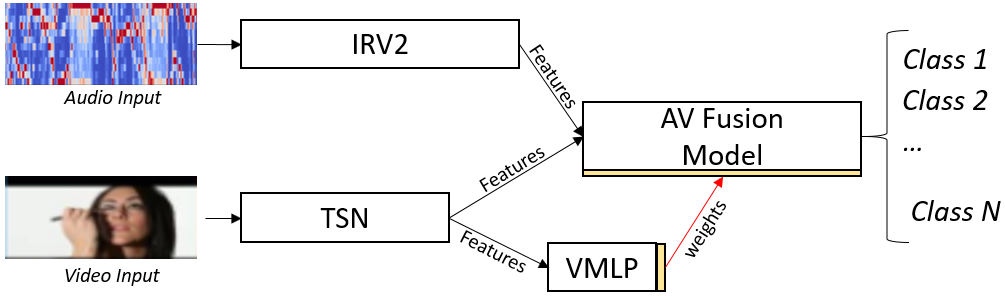}
    \caption{The proposed MAiVAR architecture.}
    \label{fig:architecture}
\end{figure}

The proposed MAiVAR architecture is illustrated in Fig.~\ref{fig:architecture}. Below, we discuss the background of our method, feature representations and network architecture.
\subsection{Background}\label{background}

\begin{figure}[b]
    \centering
    \includegraphics[width=0.45\textwidth]{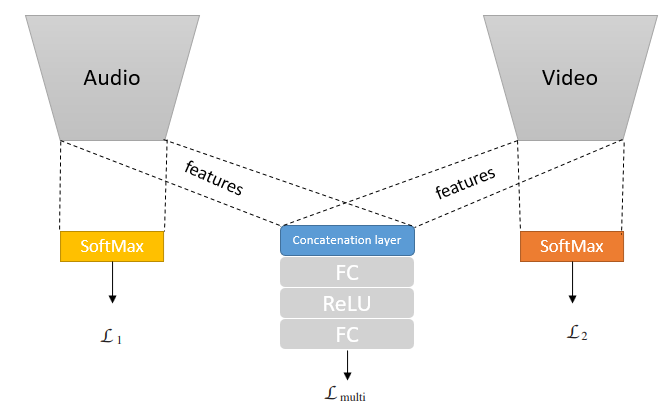}
     \caption{Multimodal architecture schematics. The audio and video models output an independent action prediction $\mathcal{L}_1$ and $\mathcal{L}_2$ based on their respective input datasets. The features from the \textit{AvgPool2d} layers of the two models are combined to feed the fusion module, which outputs the predicted action $\mathcal{L}_{multi}$.}
     \label{fig:fusion}
\end{figure}

Given the training set $T = \{{x_1...x_n, y_1...y_n}\}$,
where $x_i$ is the $i$-th training example and $y_i$
is its true label,
training on a single modality $m$ (e.g. RGB frames or audio) means minimizing an empirical loss:
\begin{equation} \label{eq:1}
L(C (\phi_m(X)), y)    
\end{equation}
where $\phi_m$ is normally a deep network with parameter $\ominus_m$,
and $C$ is a classifier, typically one or more fully-connected
(FC) layers with parameter $\ominus_c$. For classification problems
considered here, $L$ is the cross-entropy loss. Minimizing (\ref{eq:1}) gives solution $\ominus\**_m$ and $\ominus\**_c$. We train a late-fusion model on $M$ different modalities $(\{m_i\}^k_1)$. Each modality is processed by a different deep network $\phi_m{_i}$ with parameter $\ominus_m{_i}$
, with their features fused and passed to a classifier $C$. Formally, training is done by minimizing the loss: \vspace{-2mm}
\begin{equation}
\mathcal{L}_{multi} = L(C ( \phi_{audio}\oplus\phi_{video} ), y)     
\end{equation} 
where $\oplus$ denotes a fusion operation (e.g. concatenation).
Fig.~\ref{fig:fusion} shows an illustration of a joint training between $audio$ and $video$ modalities. \vspace{-2mm}

\subsection{Feature Representations}
\subsubsection{Visual Feature Representation} \label{visual}
\comment{
\begin{figure}
    \centering
    \includegraphics[width=0.45\textwidth]{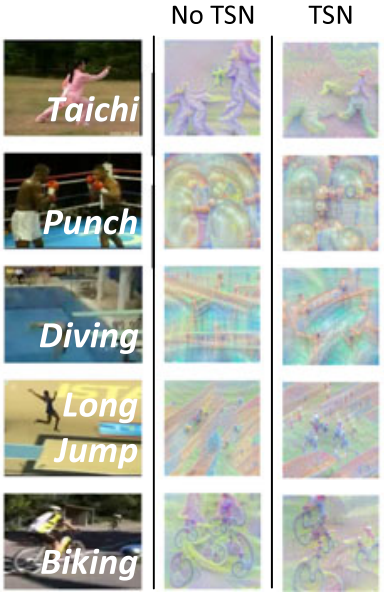}
    \caption{Visualization of ConvNet models for action recognition using DeepDraw [79]. It is worth noting that video frames are just representatives of the
corresponding classes, but not used for these visualizations. All these images are generated from purely random pixels. We compare two settings:
(1) without temporal segment network (No TSN); (2) with temporal segment network (TSN). (taken from \cite{wang2016temporal}).}
    \label{fig:ucf-tsn}
\end{figure}
}
The duration of video input is $t$ seconds, then  converted into a frame-based representation that is compatible with TSN. This is fed as input to the TSN feature extractor. Each video clip represents an action example. We extract the feature from TSN  using the \textit{AvgPool2d} layer while pruning the original average consensus layer. TSN produces embeddings of size $1024 \times 25$, to allow the model to capture the spatio-temporal structure of the RGB video. The resulting sequence is then fed as input into the fusion model. The advantages of this setup are: 1) the standard TSN architecture is easy to implement and reproduce as it is off-the-shelf in TensorFlow and PyTorch, and 2) we intend to apply transfer learning for Fusion MLP, where a standard architecture makes transfer learning easier. We have adapted the MMAction2 \cite{mmaction2} interface for using TSN as a feature extractor. TSN is used as a feature extractor based on its temporal pooling of frame-level features, which has been rigorously applied as an efficient video-level feature extractor in different problems. For example, when TSN is plugged into GSM \cite{sudhakaran2020gate}, a growth of 32\% accuracy is gained. Yang et al. \cite{yang2022sta} has used TSN with soft attention mechanism to capture important frames from each segment, where these frames are key candidates to volume-based features. Zhang  \cite{zhang2020automated} has used the TSN model as a feature extractor with ResNet101 for efficient automated pigs behavior recognition. 
\subsubsection{Audio Feature Representation} \label{audio}
The audio input has been transformed into different image-based representations. This results in $(224\times224\times3)\times n$ ($n$ is the total number of audio action instances) audio-image as input to the IRV2 backbone. Accordingly, each audio-image represents a whole action instance (equivalent to video clip of $t$ secs). We extract the feature from IRV2 using the \textit{AvgPool2d} layer while pruning the original classification layer. IRV2 produces embeddings of size $1536$ to allow the model to capture spectral information. The resulting sequence is then fed as input to the fusion model. The advantage of this simple setup is that the standard IRV2 architecture performs better on audio-image representations as compared to other CNN-based models (see Section \ref{ablation}).\vspace{-2mm}
\subsection{Network Architecture}
\subsubsection{Visual Network} \label{visualnetwork}
For simplicity, we adopt an established CNN-based visual feature extractor TSN \cite{wang2016temporal} which projects visual embeddings to the fusion model. In TSN, the original BNInception backbone is used. The consensus layer is removed to expose features from \textit{AvgPool2D} layer of the TSN model. The features from TSN are fed as input to the visual MLP. The weights from the trained visual network are then used to initialize the Fusion MLP. 
\subsubsection{Audio Network} \label{audionetwork}
The audio network  uses an IRV2-based feature extractor initialized with weights from ImageNet for creating audio embeddings for the audio MLP. In IRV2, we froze all gradients and removed the last layer to expose features from the average pooling layers of the model. The features from IRV2 are fed as input to the fusion MLP. 
\subsubsection{Fusion Network} \label{fusionnetwork}
The fusion network concatenates the audio and visual features and classifies the combined feature  into action labels. As discussed in Section \ref{visualnetwork}, the visual network produces higher classification accuracy. Therefore, weights from the trained visual network were used to initialize the fusion model to fine-tune the fusion model. \par
All hidden layers, except the last fully-connected layer, are equipped with Rectified Linear Unit (ReLU) non-linearity. The fusion network was trained by minimizing the cross-entropy loss $L$ with $l_1$ regularization using back-propagation: \begin{equation}
 \arg \min_{W}   \sum_{i,j} L (x^i_j+z^i_j,y^i_j,W)
\end{equation} where $x^i_j$ and $z^i_j$ is the $j$-th input vector from the respective audio and video features, $y_j$ is the corresponding class label and $W$ is the set of network parameters. For the fusion model, we used the initialization weights from the video classification model (see Algorithm \ref{algo1}). Empirically, the weights from the video model boost the performance of the fusion model. 
\begin{algorithm}
\small
\caption{\small Weights assignment algorithm}\label{alg:cap}
\label{algo1}
\begin{algorithmic}[1]
\Require Initial weights $\phi^0$, $ \rho^0$ for \text{audio} and \text{video} respectively, 
$N$  Number of epochs
\Ensure $\psi^0$ \newline
        \hspace*{5em}\Comment{Weights after training} 
\Procedure{Weights}{$\phi^0,\rho^0, N $}
\State Train $\phi^0 \text{with} \{Wa_i\}^k_{i=1}$ 
\For{\texttt{$i = 1,2,...,N$}}
        \State \text{process batch}
        \State \text{update weights} $\phi_i $
      \EndFor
\State  Train $\rho^0$ {with} $\{Wv_i\}^k_{i=1}$
\For{\texttt{$i = 1,2,...,N$}}
        \State \text{process batch}
        \State \text{update weights} $\rho_i $
      \EndFor
\State $ \psi^0 \gets \rho^N$;
\State Fusion model weights $\psi^0$ initialized with weights from trained video model $\rho^N$.
\State \textbf{return} $\psi^0$\Comment{Weights for fusion model}
\EndProcedure
\end{algorithmic}
\end{algorithm}

\section{EXPERIMENTAL SETUP AND RESULTS}
\label{sec:experiment}
Here we discuss the dataset, implementation details, our comparison with state-of-the-art models and the ablation study.
\subsection{Dataset}\label{dataset}
We conducted experiments on UCF101 dataset \cite{soomro2012ucf101} which is a standard benchmark for action recognition. UCF-101 contains $13,320$ videos of $101$ human action
categories, such as Apply Eye Makeup, Blow Dry Hair and Table Tennis with an average length of $180$ frames per video. We observed that half of the dataset had no audio channel. Thus, to focus on the effect of audio features, we used the subset with audio channel \cite{takahashi2017aenet}. This resulted in $6837$ videos across $51$ categories. Whilst this led the dataset to be significantly reduced,  the distribution of the audio dataset was similar to the video dataset. We used the first train-test split setting provided with this dataset, which resulted in $4893$ training and $1944$ testing samples.
\subsection{Implementation Details}\label{implement}

We have considered two input modalities: video and audio. For video, we used RGB frames as input with $25$ as the number of segments for TSN. We followed \cite{wang2016temporal} for visual pre-processing and data augmentation. For audio, we used Librosa \cite{mcfee2015librosa} to generate audio-image representations. Audio-image representations were normalized as per ImageNet configuration, with random horizontal and vertical flips. For video, \cite{wang2016temporal} was followed.  For feature extraction, the TSN-based \cite{wang2016temporal} feature extractor used the BNInception backbone for visual data and IRV2 \cite{szegedy2016inceptionv4}, using residual inception blocks for audio feature extraction.  \par
For training and testing, all networks were implemented on a machine with an Intel {XEON\textregistered}  CPU and $8\times$ GeForce GTX $1080$ Ti GPUs. The models were implemented using the PyTorch \cite{NEURIPS2019_9015}. Similar to \cite{takahashi2017aenet}, the data was randomly split as per split $1$ on UCF-101 \cite{soomro2012ucf101}. We use Adam \cite{kingma2015adam} optimizer, where the optimal choices for learning rates $\alpha$ were $3\mathrm{e}^{-4}$ for audio and video models, while for the fusion model $1\mathrm{e}^{-4}$ a batch size of $16$ and $128$ for audio-image representation and the fusion model was used, respectively.
\subsection{Comparing MAiVAR Against Other Methods on UCF-101} \label{results}
\begin{table}[t]
    \centering
    \caption{Comparison of the performance of MAiVAR with the state-of-the-art models on UCF-101 dataset.}
\begin{center}
\begin{tabular}{c c}
\toprule
\textbf{Method}        & \textbf{Accuracy} (\%) \\[0.5ex] 
\midrule
TSN (RGB) \cite{wang2016temporal}                & 60.77    \\
 IMGAUD2VID \cite{gao2020listentolook}                & 81.10 \\
C3D \cite{tran2015learning}                      & 82.23              \\
STA-Net (RGB)\cite{yang2022sta} & 83.42      \\
C3D+AENet \cite{takahashi2017aenet}                 & 85.33             \\
\midrule
%TSN (RGB + RGB Difference) & 83.8              \\ 
%\textbf{MAiVAR-WP}             & \textbf{86.21}   \\
%\textbf{MAiVAR-SC}             & \textbf{86.26}   \\
%\textbf{MAiVAR-SR}             & \textbf{86.00}    \\
%MaiVAR-MFCC             & {83.95}    \\
%\textbf{MAiVAR-MFS}             & \textbf{86.11}    \\
\textbf{MAiVAR-CH}             & \textbf{87.91}    \\
\bottomrule
\end{tabular}
\end{center}
    \label{sota}
\end{table}
We show a comparison of the performance of our model against state-of-the-art methods in Table \ref{sota}. The benchmarks were reproduced using accuracy over the standard train and test split. Following the evaluation protocol of \cite{takahashi2017aenet}, we used the accuracy to evaluate the performance of the models.
\subsection{Ablation Experiments} \label{ablation}
\begin{figure}
    \centering
    \includegraphics[width=0.30\textwidth]{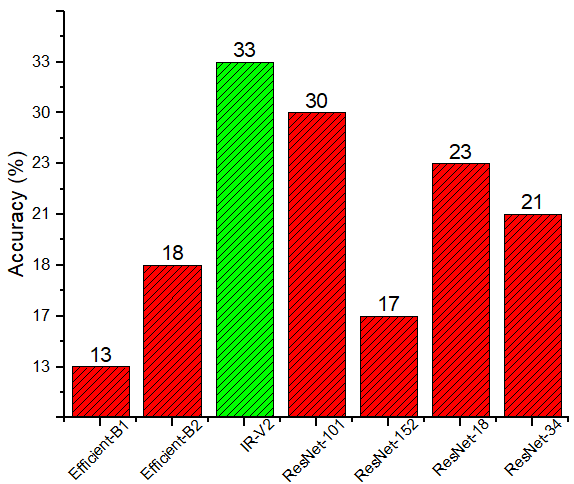}
    \caption{Comparison of audio-image only feature-extractors. Green bar highlights the best performing model.}
    \label{fig:ablation1}
\end{figure}
\subsubsection{Audio-image Feature Extraction} \label{ablation1}
Several experiments were conducted to determine the best settings for the proposed model, which can be broken down into selecting feature extractors for audio-image representations and choosing optimal audio-image representation and hyper-parameter settings. We tried several CNN-based feature extractors for audio-image representations including smaller as well as deeper variants of ResNet \cite{he2016deep}, EfficientNet \cite{tan2019efficientnet} and InceptionResNet \cite{szegedy2016inceptionv4}. Compared to IRV2, ResNet-101 and ResNet-18 showed relatively better performance (see Fig. \ref{fig:ablation1}).  
\subsubsection{Convergence of Audio Representations}
\begin{figure} %[b!]
\centering
    
    \begin{subfigure}[b]{0.12\textwidth}
         \centering
         \includegraphics[width=\textwidth]{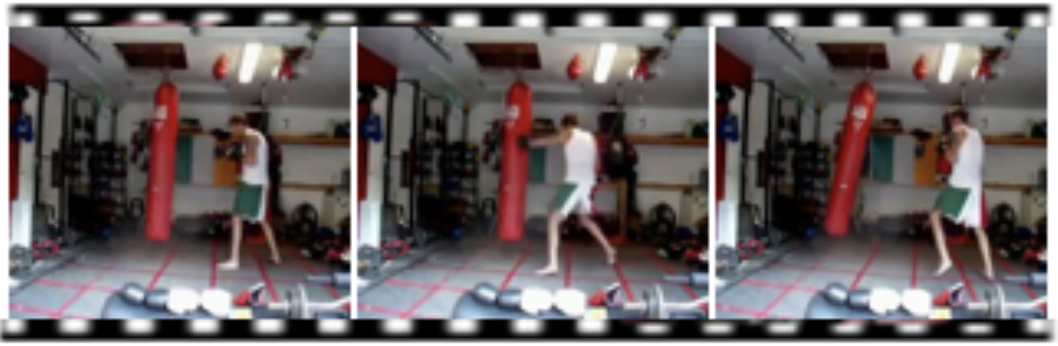}
    \caption{}
         \label{a}
     \end{subfigure}\hfil
     \begin{subfigure}[b]{0.12\textwidth}
         \centering
         \includegraphics[width=\textwidth]{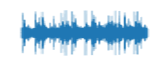}
    \caption{}
         \label{b}
     \end{subfigure}\hfil
     \begin{subfigure}[b]{0.12\textwidth}
         \centering
         \includegraphics[width=\textwidth]{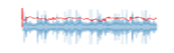}
    \caption{}
         \label{c}
     \end{subfigure}\hfil
    \begin{subfigure}[b]{0.12\textwidth}
     \centering
     \includegraphics[width=\textwidth]{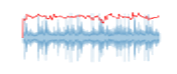}
    \caption{}
     \label{d}
    \end{subfigure}\hfil
    \vspace*{0.25cm}
    \begin{subfigure}[b]{0.12\textwidth}
     \centering
     \includegraphics[width=\textwidth]{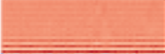}
    \caption{}
     \label{e}
    \end{subfigure}\hfil
    \begin{subfigure}[b]{0.12\textwidth}
     \centering
     \includegraphics[width=\textwidth]{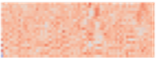}
    \caption{}
     \label{f}
    \end{subfigure}\hfil
    \begin{subfigure}[b]{0.12\textwidth}
     \centering
     \includegraphics[width=\textwidth]{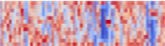}
    \caption{}
     \label{g}
    \end{subfigure}
\caption{(a) Segmented video input and six different audio-image representations of the same action: (b) Waveplot, (c) 
MFCC, (d) MFCC feature scaling, (e) Spectral Centroids,
(f) Spectral Rolloff and (g) Chromagram.}
    \label{fig:input-reps}
\end{figure}

\begin{figure}[th!]
    \centering
    \includegraphics[width=0.45\textwidth]{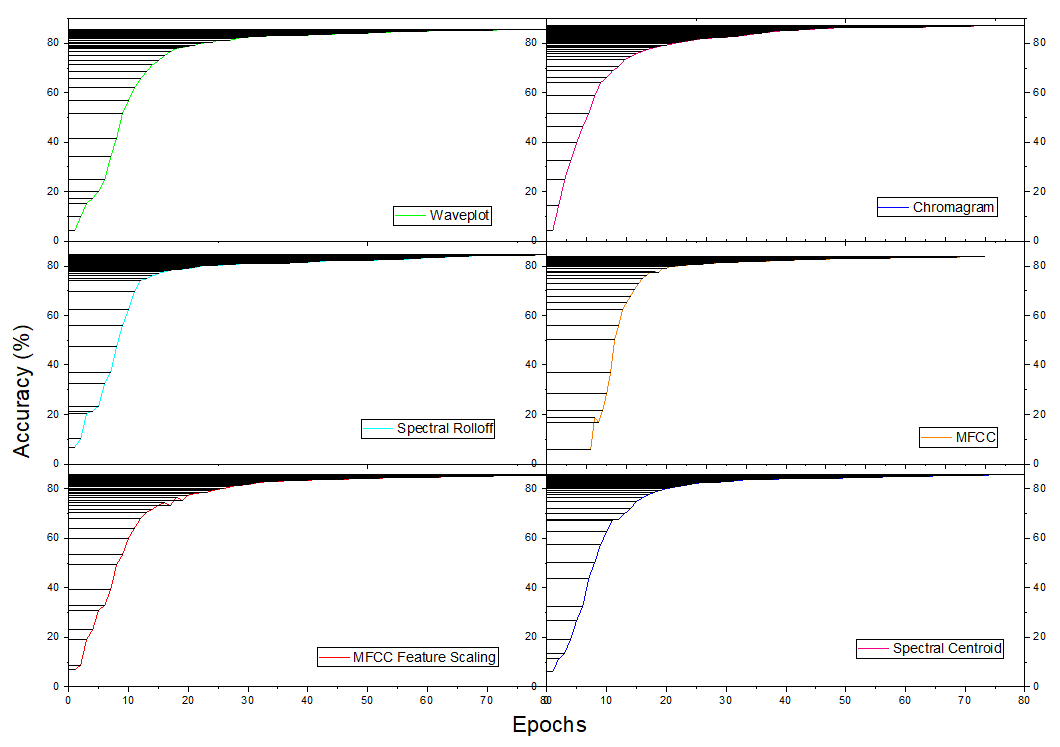}
 \caption{Convergence of the audio representations.}    \label{fig:ablation}
\end{figure}

We evaluated performance of  six different audio-image representations (shown in Fig. \ref{fig:input-reps}) for fusion with video features. Before fusion, as the structure of the chromagram is well-suited for CNN-based models, chromagram-based representations performed better than other competitor representations. Convergence of each audio-image representation after fusion with video modality is illustrated in Fig. \ref{fig:ablation}. However, it has been observed that representation that produces better accuracy within unimodal learning is not the optimal candidate for achieving better accuracy after fusion. As presented in Table \ref{audio-ablation}, MFCCs Feature Scaling-based representation produced the best audio-only accuracy, but after fusion with video, the chromagram-based representation produced the best accuracy.   

\begin{table}
    \centering
    \caption{Comparing audio-image representations before (audio MLP) and after fusion on the basis of accuracy in percentage.   \textit{Note: video-only accuracy is 75.67\% }. }
\begin{center}
\begin{tabular}{c c c}
\toprule
\textbf{Representation}           & \textbf{Audio}  & \textbf{Fusion (Audio+Video)}   \\
\midrule
Waveplot                 & 12.08 &   86.21          \\
Spectral Centroids        & 13.22       & 86.26      \\      
Spectral Rolloff              & 16.46 &  86.00           \\
MFCCs                & 12.96     &  83.95        \\
MFCCs Feature Scaling                & \textbf{17.43}     &   86.11       \\
Chromagram                        & 15.48  &  \textbf{87.91} \\       
\bottomrule
%TSN (RGB + RGB Difference) & 83.8              \\ 
\end{tabular}
\end{center}
    \label{audio-ablation}
\end{table}

\section{Conclusion}
\label{sec:conclusion}
We propose MAiVAR, a fusion-based, end-to-end model for audio-video classification that features both simple architecture and superior performance. The experimental results show that our audio-image to video fusion-based model performs well in comparison with state-of-the-art models. The fusion module is significant in extracting joint information among different modalities. This is because it is evident that pre-training on a larger video dataset improves results substantially. Our framework can be applied to large-scale multimodal action recognition datasets, such as Kinetics $400/600/700$ \cite{smaira2020short}.
\section*{Acknowledgments}
This work is jointly supported by Edith Cowan University (ECU) and Higher Education Commission (HEC) of Pakistan under Project \#PM/HRDI-UESTPs/UETs-I/Phase-1/Batch-VI/2018. Dr. Akhtar is a recipient of Office of National Intelligence National Intelligence Postdoctoral Grant \# NIPG-2021-001 funded by the Australian Government.
%%%%%%%%% BODY TEXT

\bibliographystyle{IEEEtran}
\bibliography{main}

\begin{thebibliography}{10}
\providecommand{\url}[1]{#1}
\csname url@samestyle\endcsname
\providecommand{\newblock}{\relax}
\providecommand{\bibinfo}[2]{#2}
\providecommand{\BIBentrySTDinterwordspacing}{\spaceskip=0pt\relax}
\providecommand{\BIBentryALTinterwordstretchfactor}{4}
\providecommand{\BIBentryALTinterwordspacing}{\spaceskip=\fontdimen2\font plus
\BIBentryALTinterwordstretchfactor\fontdimen3\font minus
  \fontdimen4\font\relax}
\providecommand{\BIBforeignlanguage}[2]{{%
\expandafter\ifx\csname l@#1\endcsname\relax
\typeout{** WARNING: IEEEtran.bst: No hyphenation pattern has been}%
\typeout{** loaded for the language `#1'. Using the pattern for}%
\typeout{** the default language instead.}%
\else
\language=\csname l@#1\endcsname
\fi
#2}}
\providecommand{\BIBdecl}{\relax}
\BIBdecl

\bibitem{shaikh2021rgb}
M.~B. Shaikh and D.~Chai, ``{RGB-D data-based action recognition}:{ a
  review},'' \emph{Sensors}, vol.~21, no.~12, p. 4246, 2021.

\bibitem{li2019collaborative}
C.~Li \emph{et~al.}, ``Collaborative spatiotemporal feature learning for video
  action recognition,'' in \emph{Proc. of the CVPR}, 2019, pp. 7872--7881.

\bibitem{wu2018compressed}
C.-Y. Wu \emph{et~al.}, ``Compressed video action recognition,'' in \emph{Proc.
  of the CVPR}, 2018, pp. 6026--6035.

\bibitem{wang2016temporal}
L.~Wang \emph{et~al.}, ``Temporal segment networks: Towards good practices for
  deep action recognition,'' in \emph{Proc. of the {ECCV}}, 2016, pp. 20--36.

\bibitem{zhou2018temporal}
B.~Zhou \emph{et~al.}, ``Temporal relational reasoning in videos,'' in
  \emph{Proc. of the {ECCV}}, 2018, pp. 803--818.

\bibitem{lin2019tsm}
J.~Lin, C.~Gan, and S.~Han, ``{TSM}: Temporal shift module for efficient video
  understanding,'' in \emph{Proc. of the ICCV}, 2019, pp. 7083--7093.

\bibitem{ren2015faster}
S.~Ren \emph{et~al.}, ``Faster {R-CNN}: Towards real-time object detection with
  region proposal networks,'' in \emph{Proc. of the NeurIPS}, vol.~28, 2015,
  pp. 91--99.

\bibitem{russakovsky2015imagenet}
O.~Russakovsky \emph{et~al.}, ``{ImageNet} large scale visual recognition
  challenge,'' \emph{Int. Journal of Computer Vision}, vol. 115, no.~3, pp.
  211--252, 2015.

\bibitem{gaver1993world}
W.~W. Gaver, ``What in the world do we hear?: An ecological approach to
  auditory event perception,'' \emph{Ecological psychology}, vol.~5, no.~1, pp.
  1--29, 1993.

\bibitem{kazakos2019epicfusion}
E.~Kazakos \emph{et~al.}, ``{EPIC-Fusion}: Audio-visual temporal binding for
  egocentric action recognition,'' in \emph{Proc. of the ICCV}, 2019, pp.
  5492--5501.

\bibitem{feichtenhofer2019slowfast}
C.~Feichtenhofer \emph{et~al.}, ``Slowfast networks for video recognition,'' in
  \emph{Proc. of the {ICCV}}, 2019, pp. 6202--6211.

\bibitem{carreira2017quo}
J.~Carreira and A.~Zisserman, ``{Quo Vadis}, action recognition? a new model
  and the kinetics dataset,'' in \emph{Proc. of the CVPR}, 2017, pp.
  6299--6308.

\bibitem{szegedy2016inceptionv4}
C.~Szegedy \emph{et~al.}, ``Inception-v4, inception-resnet and the impact of
  residual connections on learning,'' in \emph{Proc. of the AAAI}, 2017, pp.
  4278--4284.

\bibitem{Lei_2021_CVPR}
J.~Lei \emph{et~al.}, ``Less is more: Clipbert for video-and-language learning
  via sparse sampling,'' in \emph{Proc. of the CVPR}, June 2021, pp.
  7331--7341.

\bibitem{girdhar2017actionvlad}
R.~Girdhar \emph{et~al.}, ``{ActionVLAD}: Learning spatio-temporal aggregation
  for action classification,'' in \emph{Proc. of the CVPR}, 2017, pp. 971--980.

\bibitem{li2016vlad3}
Y.~Li \emph{et~al.}, ``{VLAD3}: Encoding dynamics of deep features for action
  recognition,'' in \emph{Proc. of the CVPR}, 2016, pp. 1951--1960.

\bibitem{Kwon_2021_ICCV}
H.~Kwon \emph{et~al.}, ``Learning self-similarity in space and time as
  generalized motion for video action recognition,'' in \emph{Proc. of the
  ICCV}, Oct. 2021, pp. 13\,065--13\,075.

\bibitem{mei2020artificial}
X.~Mei \emph{et~al.}, ``Artificial intelligence--enabled rapid diagnosis of
  patients with {COVID-19},'' \emph{Nature Medicine}, vol.~26, no.~8, pp.
  1224--1228, 2020.

\bibitem{gu2021ntire}
J.~Gu \emph{et~al.}, ``{NTIRE} 2021 challenge on perceptual image quality
  assessment,'' in \emph{Proc. of the CVPR}, 2021, pp. 677--690.

\bibitem{yan2021precise}
C.~Yan \emph{et~al.}, ``Precise no-reference image quality evaluation based on
  distortion identification,'' \emph{ACM TOMM}, vol.~17, no.~3s, pp. 1--21,
  2021.

\bibitem{he2019stnet}
D.~He \emph{et~al.}, ``Stnet: Local and global spatial-temporal modeling for
  action recognition,'' in \emph{Proc. of the AAAI}, vol.~33, no.~01, 2019, pp.
  8401--8408.

\bibitem{takahashi2017aenet}
N.~Takahashi, M.~Gygli, and L.~Van~Gool, ``{AENet}: Learning deep audio
  features for video analysis,'' \emph{IEEE TMM}, vol.~20, no.~3, pp. 513--524,
  2017.

\bibitem{gao2020listentolook}
R.~Gao \emph{et~al.}, ``Listen to look: Action recognition by previewing
  audio,'' in \emph{Proc. of the CVPR}, 2020, pp. 10\,457--10\,467.

\bibitem{mmaction2}
\BIBentryALTinterwordspacing
``Welcome to mmaction2's documentation!'' [Online]. Available:
  \url{https://mmaction2.readthedocs.io/en/latest/}
\BIBentrySTDinterwordspacing

\bibitem{sudhakaran2020gate}
S.~Sudhakaran, S.~Escalera, and O.~Lanz, ``Gate-shift networks for video action
  recognition,'' in \emph{Proc. of the CVPR}, 2020, pp. 1102--1111.

\bibitem{yang2022sta}
G.~Yang \emph{et~al.}, ``{STA-TSN}: Spatial-temporal attention temporal segment
  network for action recognition in video,'' \emph{PloS one}, vol.~17, no.~3,
  pp. 1--19, 2022.

\bibitem{zhang2020automated}
K.~Zhang \emph{et~al.}, ``Automated video behavior recognition of pigs using
  two-stream convolutional networks,'' \emph{Sensors}, vol.~20, no.~4, p. 1085,
  2020.

\bibitem{soomro2012ucf101}
K.~Soomro, A.~R. Zamir, and M.~Shah, ``{UCF101}: A dataset of 101 human actions
  classes from videos in the wild,'' \emph{arXiv preprint arXiv:1212.0402},
  2012.

\bibitem{mcfee2015librosa}
B.~McFee \emph{et~al.}, ``{Librosa}: Audio and music signal analysis in
  python,'' in \emph{Proc. of the Python in Science Conf.}, vol.~8, 2015, pp.
  18--25.

\bibitem{NEURIPS2019_9015}
A.~Paszke \emph{et~al.}, ``{PyTorch}: An imperative style, high-performance
  deep learning library,'' in \emph{Proc. of the NeurIPS}, 2019, pp.
  8024--8035.

\bibitem{kingma2015adam}
\BIBentryALTinterwordspacing
D.~P. Kingma and J.~Ba, ``Adam: A method for stochastic optimization,'' in
  \emph{ICLR (Poster)}, 2015. [Online]. Available:
  \url{http://arxiv.org/abs/1412.6980}
\BIBentrySTDinterwordspacing

\bibitem{tran2015learning}
D.~Tran \emph{et~al.}, ``Learning spatiotemporal features with 3d convolutional
  networks,'' in \emph{Proc. of the ICCV}, 2015, pp. 4489--4497.

\bibitem{he2016deep}
K.~He \emph{et~al.}, ``Deep residual learning for image recognition,'' in
  \emph{Proc. of the CVPR}, 2016, pp. 770--778.

\bibitem{tan2019efficientnet}
M.~Tan and Q.~Le, ``Efficientnet: Rethinking model scaling for convolutional
  neural networks,'' in \emph{ICML}, 2019, pp. 6105--6114.

\bibitem{smaira2020short}
L.~Smaira \emph{et~al.}, ``A short note on the kinetics-700-2020 human action
  dataset,'' \emph{arXiv preprint arXiv:2010.10864}, 2020.

\end{thebibliography}

\end{document}